# ArPanEmo: An Open-Source Dataset for Fine-Grained Emotion Recognition in Arabic Online Content during COVID-19 Pandemic

Maha Jarallah Althobaiti
Department of Computer Science
College of Computers and Information Technology
Taif University
P.O.BOX 11099, Taif 21944, Saudi Arabia
Maha.j@tu.edu.sa

**Abstract**
Emotion recognition is a crucial task in Natural Language Processing (NLP) that enables machines to comprehend the feelings conveyed in the text. The task involves detecting and recognizing various human emotions like anger, fear, joy, and sadness. The applications of emotion recognition are diverse, including mental health diagnosis, student support, and the detection of online suspicious behavior. Despite the substantial amount of literature available on emotion recognition in various languages, Arabic emotion recognition has received relatively little attention, leading to a scarcity of emotion-annotated corpora. This paper presents the ArPanEmo dataset, a novel dataset for fine-grained emotion recognition of online posts in Arabic. The dataset comprises 11,128 online posts manually labeled for ten emotion categories or neutral, with Fleiss' kappa of 0.71. It is unique in that it focuses on the Saudi dialect and addresses topics related to the COVID-19 pandemic, making it the first and largest of its kind. Python's packages were utilized to collect online posts related to the COVID-19 pandemic from three sources: Twitter, YouTube, and online newspaper comments between March 2020 and March 2022. Upon collection of the online posts, each one underwent a semi-automatic classification process using a lexicon of emotion-related terms to determine whether it belonged to the neutral or emotional category. Subsequently, manual labeling was conducted to further categorize the emotional data into fine-grained emotion categories. We anticipate that the ArPanEmo dataset will enrich Arabic NLP resources and help in the development of machine learning and deep learning tools to identify emotions in a given text. It will also contribute to developing systems that monitor online suspicious behaviors or mental health disorders. The final dataset is formatted in CSV, consisting of three columns: the number of the post, the post's text, and the corresponding emotion label. This format facilitates incorporating and utilizing the dataset in any machine learning research.

**Specifications table**

| | |
|---|---|
| **Subject** | Artificial Intelligence |
| **Specific subject area** | Arabic Natural Language Processing, Emotion Recognition in Arabic language. |
| **Type of data** | Text (.csv files) |
| **How the data were acquired** | Using Python's packages, namely, Tweepy, BeautifulSoup, Lxml, and Urllib. |
| **Data format** | Raw |
| **Description of data collection** | We collected online posts related to the COVID-19 pandemic from three sources: Twitter (using Tweepy package), YouTube comments (using Urllib and Lxml packages), and online newspaper comments (using BeautifulSoup package). Our focus was on topics related to the pandemic, such as social distancing, sanitization, and emerging Coronavirus-related apps, as well as the impact of COVID-19 on various aspects of life. The collected online posts were manually annotated for ten emotion categories or neutral. |
| **Data source location** | Middle East, Saudi Arabia |
| **Data accessibility** | Repository name: Mendeley data<br><br>Data identification number: 10.17632/d9yy8w52ns.2<br><br>Direct URL to data: https://data.mendeley.com/datasets/d9yy8w52ns. |

**Value of the data**
- Due to COVID-19, social media platforms have become popular channels for people to express their emotions and exchange knowledge. The significant influence of emotions on individuals and groups' behavior during a pandemic highlights the importance of identifying these emotions and understanding their sources. The ArPanEmo dataset can be beneficial in developing automated systems to monitor behaviors and mental health disorders during such crises.
- The lack of annotated datasets is impeding progress in Arabic emotion recognition research. Creating such a dataset could enhance both Arabic NLP overall and Arabic emotion recognition specifically.
- Apart from the insufficient number of emotion-annotated datasets, current datasets for Arabic emotion recognition cover general topics, with a shortage of specialized datasets focusing on specific topics. Therefore, the ArPanEmo dataset, as a specialized resource, can facilitate further investigation in future research on topic-specific emotion recognition.
- The Arabic language comprises various Arabic dialects and Modern Standard Arabic (MSA). Existing datasets mainly consist of instances written in different Arabic dialects and MSA. However, the ArPanEmo dataset focuses on one specific Arabic variety, namely Saudi Arabic, which can support future research on specific Arabic dialects.
- The ArPanEmo dataset can be used to develop machine learning and deep learning models, enabling the identification of emotions from any given text.
- The data collection methodology described can be replicated to produce datasets for various Arabic varieties or languages like English. Moreover, the methodology can be utilized to collect data for additional emotion categories beyond those included in the ArPanEmo dataset.

## 1. Objective

Emotion recognition is a vital task in NLP which has many applications in various fields including healthcare, education, and security. Nevertheless, Arabic emotion recognition has been under-addressed as a result of the limited number of emotion-annotated corpora. This paper presents the ArPanEmo dataset that we have collected from online posts and manually annotated for ten emotion categories or neutral. The main objective of this dataset is to enrich Arabic NLP resources and contribute to Arabic emotion recognition. As this dataset is collected from real-world online content, it can be used to train machine learning/deep learning models for real-world applications that monitor emotional states of online users, support teachers to understand the students' needs, or track online suspicious behaviors.

## 2. Data description

The data collected focuses on topics related to the pandemic, representing two main groups: healthcare and Impacted Life Aspects (ILA) by COVID-19. The healthcare domain included topics related to the diagnosis, treatment, and prevention of COVID-19, while the ILA domain covered topics such as school closures, remote work, emerging Coronavirus-related apps, and stock market changes.

The ArPanEmo dataset consists of 11,128 online posts, each labeled with one of 11 categories: anger, anticipation, confusion, disgust, fear, joy, neutral, optimism, pessimism, sadness, and surprise. In total, the dataset contains 223,008 tokens (i.e., words). Within the dataset, there are 8,751 posts related to healthcare and 2,377 posts related to ILA domain.

In order to obtain additional information about the distribution of the length of the online posts in ArPanEmo dataset, we compute the post's length, in terms of the number of characters and tokens, their mean, standard deviation, and a five-number summary. Table 1 presents the statistical summaries for text length in ArPanEmo. The average number of characters in the post is about 112. There is a minimum of three characters in a post. In terms of the statistical information about the tokens in the posts, ArPanEmo has an average length of around 19 tokens and a maximum length of 127 tokens. Posts with less than 10 tokens make up 25% of the dataset while 75% of the posts contain less than 25 tokens.

**Table 1.** Statistical summaries for text length in ArPanEmo dataset.

|  | mean | std | min | 25% | 50% | 75% | max |
|---|---|---|---|---|---|---|---|
| **Text length (Char)** | 111.91 | 70.61 | 3.00 | 59.00 | 92.00 | 148.00 | 709.00 |
| **Text length (Token)** | 19.24 | 12.10 | 1.00 | 10.00 | 16.00 | 25.00 | 127.00 |

The aim of the development of ArPanEmo dataset is to build models for emotion recognition using machine learning or deep learning techniques. The implementation process requires training the models and then evaluate their performances using test set. Consequently, in order to ensure reproducibility when constructing models, the training and test sets of the ArPanEmo dataset have been clearly defined. The dataset has been split into two sections: the training set, which accounts for 80% of the data, and the test set, which accounts for 20%. The division was conducted randomly. Nevertheless, to balance the training and test sets across different emotion categories and domains, we divided the ArPanEmo dataset based on the number of instances in each category and domain. Specifically, we ensured that 80% of the posts in the corpus from each domain (healthcare and ILA) were included in the training set, while the remaining 20% were assigned to the test set. We followed the same procedure for each emotion category. See Table 2 for further details regarding the distribution of ArPanEmo training and test sets across emotion categories and domains.

**Table 2.** The distribution of ArPanEmo training and test sets.

| Class | Train (80%) | | Test (20%) | |
|---|---|---|---|---|
|  | **Health** | **ILA** | **Health** | **ILA** |
| anger | 863 | 262 | 216 | 66 |
| anticipation | 478 | 154 | 119 | 39 |
| confusion | 642 | 99 | 160 | 25 |
| disgust | 500 | 65 | 126 | 16 |
| fear | 880 | 113 | 220 | 28 |
| happiness | 582 | 166 | 145 | 42 |
| neutral | 637 | 656 | 159 | 164 |
| optimism | 698 | 162 | 175 | 41 |
| pessimism | 574 | 75 | 143 | 19 |
| sadness | 637 | 86 | 159 | 21 |
| surprise | 510 | 62 | 128 | 16 |
| **Total** | 8,901 | | 2,227 | |

The ArPanEmo dataset contains two files. The first file is the training set of the ArPanEmo, containing 8,901 records while the second file is the test set with 2,227 records. Each record represents one online post and consists of three columns:

- **Number:** Twitter ID (for tweets) or 0 (for YouTube and online newspaper comments).
- **Post:** the post's text
- **Label:** the corresponding emotion label for each post.

A glimpse of ArPanEmo has been shown in Figure 1 with English translation.

| Number | Post | Label |
|---|---|---|
| 0 | الي مستانسين حاولو تحسون شوي 😠<br>Those who are happy, try to feel a little 😠 | anger |
| 1374297130106511368 | جتني أم الركب يوم حجزت لقاح كورونا اليوم<br>*Âum Alrkb* came to me when I booked the Corona vaccine today<br>("*Âum Alrkb*" phrase describes the state of extreme panic and fear) | fear |
| 0 | الحمدلله ربي شفاني من كورونا ونزلت لاولادي من اسبوع وحاليا تحسنت تماما الحمدلله<br>God healed me from Corona, I went to my children a week ago, and now I have fully recovered, thank God. | happiness |

**Figure 1.** Sample of the ArPanEmo dataset.

| | Main keywords (mandatory) | | | |
|---|---|---|---|---|
| | \"coronavirus\" كارونا/كرونا/كورنا/كورونا | | \"COVID-19\" كوفيد-19/كوفيد | |
| | \"pandemic\" جائحة | | \"epidemic\" وباء | |
| | **Domain-specific keywords (optional)** | | | |
| **Healthcare** | أعراض<br>\"symptoms\" | إجراءات<br>\"measures\" | لقاح<br>\"vaccine\" | علاج<br>\"therapy\" |
| | فحص<br>\"checkup\" | حظر التجول<br>\"lockdown\" | عزل<br>\"isolation\" | جرعة<br>\"dose\" |
| | تعقيم<br>\"disinfecting\" | عدوى<br>\"infection\" | | |
| **Impacted Life Aspects (ILA)** | دراسة<br>\"studying\" | حضوري<br>\"offline/in person\" | تعليم<br>\"learning\" | طالب<br>\"student\" |
| | أسهم<br>\"stocks\" | طالبات/طلاب<br>\"students\" | عمل<br>\"work/business\" | اختبارات<br>\"exams\" |
| | مدرسة<br>\"school\" | تعليق الدراسة<br>\"school closures\" | مبادرة<br>\"initiative\" | سوق<br>\"market\" |
| | أسهم<br>\"stocks\" | تطبيق توكلنا<br>\"Tawakkalna app\" | تطبيق صحتي<br>\"Shty app\" | عن بعد<br>\"online\" |
| | فعاليات<br>\"activities\" | تطبيق اعتمرنا<br>\"Eatmarna app\" | تطبيق تباعد<br>\"Tabaud app\" | أستاذ/معلم<br>\"teacher\" |
| | منصة<br>\"platform\" | ضرايب<br>\"taxes\" | اقتصاد<br>\"economics\" | |

**Figure 2.** Keywords used to collect the data of ArPanEmo dataset.

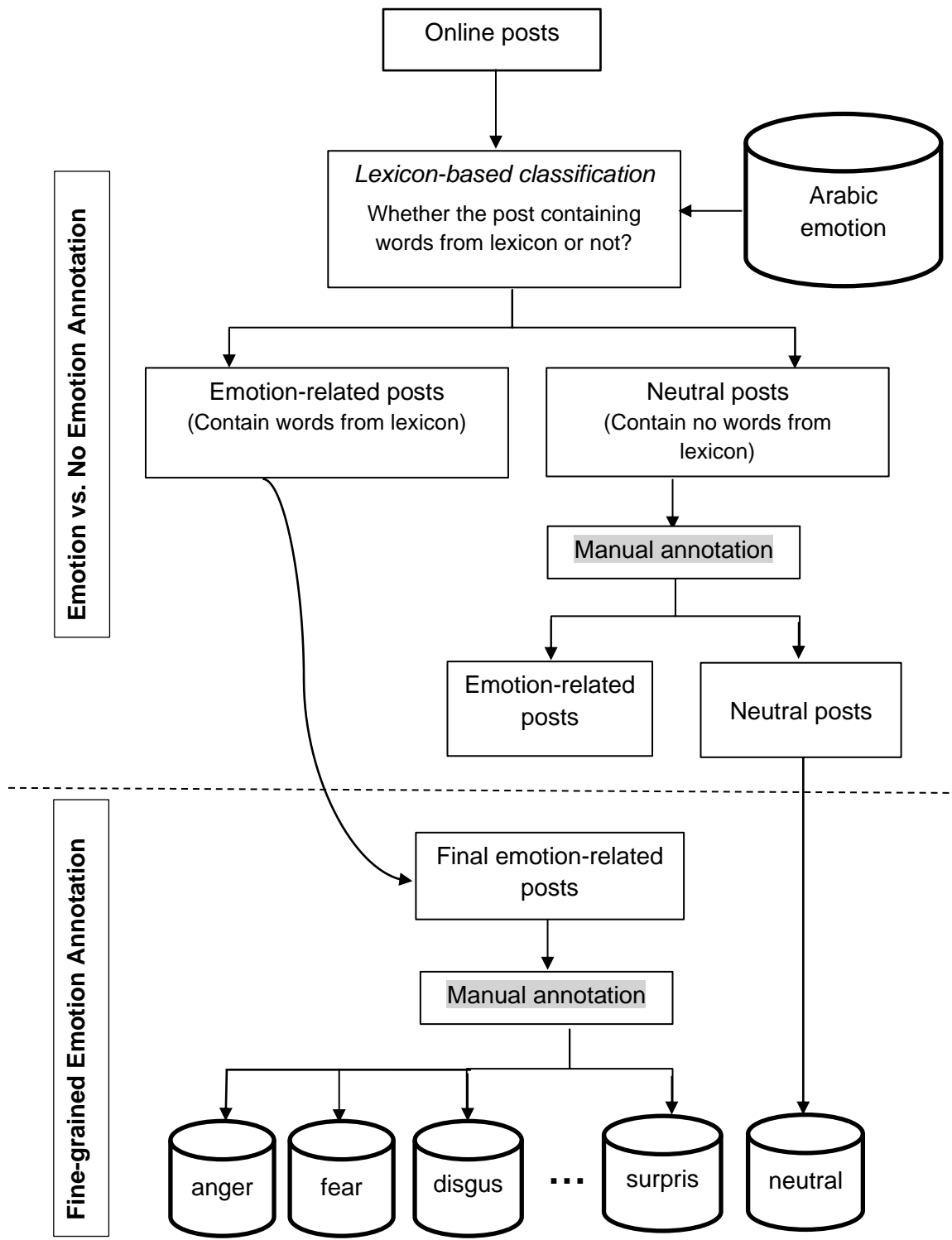

**Figure 3.** Annotation process for ArPanEmo dataset.

**Table 3.** Fleiss' Kappa for inter-annotator reliability of the ArPanEmo dataset

| Method | Coefficient | Inference/Subjects | | |
| --- | --- | --- | --- | --- |
| | | StdErr | 95% C.I. | p-Value |
| **Fleiss' Kappa** | 0.71232 | 0.04377 | 0.625 to 0.8 | 0.000E+00 |
| **Percent Agreement** | 0.74648 | 0.03868 | 0.669 to 0.824 | 0.000E+00 |

**Table 4.** Evaluation results for various emotion recognition models built based on ArPanEmo.

| Model | Accuracy | Macro-average | | | Weighted-average | | |
| --- | --- | --- | --- | --- | --- | --- | --- |
| | | F1-score | Precision | Recall | F1-score | Precision | Recall |
| **BERT-based** | 0.67 | 0.66 | 0.68 | 0.67 | 0.67 | 0.69 | 0.67 |
| **BiGRU** | 0.40 | 0.38 | 0.44 | 0.39 | 0.40 | 0.48 | 0.39 |
| **SVM** | 0.58 | 0.55 | 0.69 | 0.52 | 0.58 | 0.68 | 0.58 |

## 3. Experimental design, materials and methods

The construction of the ArPanEmo dataset involved two primary tasks: collecting data and annotating it. To ensure the consistency and reliability of the annotations, inter-annotator agreement was measured. The final ArPanEmo dataset was then validated through the development of machine learning and deep learning models for fine-grained emotion recognition.

### 3.1 Data Collection

Over a period of 24 months following the World Health Organization's declaration of COVID-19 as a pandemic on 11 March 2020, we collected data related to the virus from three sources: Twitter, YouTube, and online newspaper comments.

- Twitter: To collect Twitter data, we used the Tweedy python library [1] to access the Twitter API. We targeted users in Saudi Arabia by collecting tweets with geographical information about their location (i.e., longitude and latitude information). We used specific phrases as queries to collect pandemic-related tweets, with the main keywords shown in Figure 2. At least one main keyword had to be present in a tweet to be considered for inclusion in our dataset, while additional domain-specific keywords were optional. If a tweet contained one of the main keywords but lacked any domain-specific keyword, it was automatically considered an instance of the healthcare domain.
- YouTube: Comments were extracted from videos that had any of the main keywords listed in Figure 2 within their titles and also included a hashtag referencing Saudi Arabia (e.g., `#saudi`). The domain-specific keywords present in the video titles indicated the domain of the extracted comments. In cases where the keywords for domains were absent from video titles, we automatically assigned the collected comments to the healthcare domain. We used the lxml library [2] and urllib package [3] for processing XML and HTML in the Python.
- Online Newspapers: We conducted an assessment of various Saudi online newspapers and ultimately chose *Al-marsd*[1] due to its popularity and the significant number of comments provided by the readers. Our approach involved gathering the readers' comments on newspaper articles that featured one of the main keywords outlined in Figure 2. In addition, we applied domain-specific phrases in the same manner as we did when collecting data from Twitter and YouTube. The BeautifulSoup Python library [4] was used to collect the comments.

---
[1] https://al-marsd.com/

### 3.2 Data Annotation

To reduce cognitive burden, the annotation process was split into two sequential stages. In the first stage, the task was solely to specify if a given post conveyed any emotion. In the second stage, the task is to identify the specific emotion expressed within each post that had been marked as emotional in the previous stage. Figure 3 depicts the two stages involved in the annotation process, which will be described in detail shortly.

For the first stage, we utilized the Arabic translations of the NRC Affect Intensity Lexicon (NRC-AIL) [5, 6, 7] to filter the collected online posts and extract only those that conveyed the emotional state of the writers. The NRC-AIL contains 9,365 Arabic words that correspond to eight primary emotions in accordance with Plutchik's wheel of emotions [8]: anger, fear, disgust, sadness, surprise, trust, anticipation, and joy. The lexicon comprises translations of commonly used English emotion-related terms into Arabic, along with words that tend to co-occur with emotional terms to varying degrees. After applying the lexicon to filter the online posts, we were able to extract around 600,000 posts that conveyed emotions. We randomly selected 4,500 of these posts for annotation with one of 10 different emotions to create the ArPanEmo dataset.

After conducting a preliminary analysis of the online posts that were excluded by the NRC Affect Intensity Lexicon (NRC-AIL), we observed that numerous online posts were discarded despite containing solid words and phrases that described the emotional states of their authors. This could be due to the fact that these meaningful emotion-related words are specific to certain Arabic dialects and cannot be found in Modern Standard Arabic (MSA), which is the Arabic variety used in the NRC-AIL lexicon. That is, the lexicon only includes MSA translations of English words. We realized the importance of preserving online posts that were originally written in Arabic dialects and conveyed emotional states that were not detected by the NRC-AIL lexicon. In order to achieve this goal, we randomly chose 8,000 online posts that were identified as "No Emotion" by the NRC-AIL lexicon in the previous step. We then assigned two human annotators, who were college-educated Saudi native Arabic speakers, to each post and asked them to categorize it as either "Emotion" or "No Emotion/Neutral". The annotators followed a set of guidelines to ensure consistency in their classification process as follows:

- Emotion: The online post expresses a feeling directed toward a specific object, caused by a situation or people, such as anger, sadness, happiness, fear, ...etc.
- No Emotion/Neutral: The online post provides only knowledge absent of feeling toward a specific object.

The NRC-AIL lexicon identified 8,000 online posts as "neutral". However, after manual annotation by two human annotators, 5,318 of these posts were identified as expressing emotions. As a result, we merged the 5,318 posts labeled as "Emotion'" with the 4,500 posts identified by the lexicon.

In the second stage of annotation, the online posts were manually labeled with one of ten emotions: anger, fear, happiness, sadness, anticipation, disgust, surprise, optimism, pessimism, or confusion. We added 1,600 online posts that were considered neutral by both the lexicon and the annotator in the previous stage to the final dataset. To ensure high-quality annotations, we employed 80 college-educated Saudi native Arabic speakers to label the data, with each online post labeled by three annotators. We provided annotators with examples of online posts for each emotion category and described each emotion to ensure clarity and consistency in the labeling task as follows:
- anger: also includes frustration, annoyance, hostility
- anticipation: also includes interest, awaiting an expected event, and attention
- disgust: also includes intense dislike, hatred, revulsion
- fear: also includes anxiety, panic, fright, apprehension
- happiness: also includes joy, cheer, satisfaction, contentment, and fulfillment
- sadness: also includes grief, depression, sorrow, and gloom
- surprise: amazement following something unexpected whether it is positive, negative or neutral
- confusion: also includes uncertainty, doubts, and suspicion
- optimism: also includes hopefulness and confidence
- pessimism: also includes hopelessness and no confidence

Each post was ultimately labeled with the emotion selected by the majority of annotators; each annotator had one vote and the emotion category with the highest number of votes was chosen. When three different emotion categories were selected by the annotators, the post was neglected. At the end of the annotation process, we neglected 290 posts. In addition, 16 posts were annotated as "neutral" in the second stage. Therefore, we added them to the neutral posts.

**3.3 Inter-annotator Agreement**

In ArPanEmo, each post was assigned three annotations. However, annotating the dataset for emotion recognition was challenging due to the fact that annotators may have different opinions about the emotions expressed in the posts. Moreover, there are no hard-limit criteria to assess the emotions conveyed in short texts like online posts, making it even more difficult to annotate them accurately. In order to measure the reliability of agreement between the annotators of our ArPanEmo dataset, we used Fleiss' kappa; one of the most commonly used measures to determine agreement between three or more annotators assigning categorical ratings to a set of items [9, 10]. Fleiss' kappa coefficient, κ, is computed according to the following equation:

$$\kappa = \frac{\bar{P} - \bar{P}_e}{1 - \bar{P}_e} \qquad (1)$$

where $\bar{P}$ is the observed agreement among annotators while $\bar{P}_e$ is the hypothetical probability of chance agreement. Consequently, the factor $1 - \bar{P}_e$ gives the degree of agreement attainable above chance and the factor $\bar{P} - \bar{P}_e$ gives the degree of agreement achieved above chance. The $\kappa$ coefficient equal to 1 indicates a total agreement between annotators whereas the $\kappa$ coefficient less than or equal to zero indicates there is no agreement between the annotators. The $\bar{P}$ and $\bar{P}_e$ are calculated using the following two equations:

$$\bar{P} = \frac{1}{Nn(n-1)} \left( \sum_{i=1}^{N} \sum_{j=1}^{k} n_{ij}^2 - Nn \right) \qquad (2)$$

$$\bar{P}_e = \sum_{j=1}^{k} p_j^2 \qquad (3)$$

$$p_j = \frac{1}{Nn} \sum_{i=1}^{N} n_{ij} \qquad (4)$$

Where N represents the number of dataset instances, n represents the number of annotations per instance, and k represents the number of emotion categories. Let $n_{ij}$ represent the number of annotators who assigned the *i*-th instance to the *j*-the emotion category

The obtained value of Fleiss's kappa coefficient, $\kappa$, for our dataset which contains 10 emotion labels is equal to 0.7123.

Table *3* shows the results including the Fleiss' kappa coefficient $\kappa$, together with the 95% confidence interval (CI), and the p-value given for the test. We used AgreeStat360 software [11] for computing the measures and the standard errors.

According to Landis and Koch [12] interpretation of the kappa value, the obtained κ =0.7123 for our ArPanEmo dataset shows substantial agreement among the annotators.

**3.4 Data Validation**

Several models were trained for emotion recognition using the dataset to evaluate its effectiveness. The first step involved preprocessing the dataset to remove unnecessary characters from the raw text and normalize letters that are usually written interchangeably, which can lead to data sparsity. Preprocessing steps we applied include:
- removing URLs, mentions, retweet and hashtag symbols,
- replacing underscores in hashtag texts with spaces,
- removing all diacritical marks and punctuation,
- removing repeated letters in the words,
- removing English words,
- removing letter elongation in Arabic,
- normalizing different forms of Arabic letters.

For building machine learning model, we used Support Vector Machine (SVM). As features to train the model, we used word n-grams with *n* in the range (1,3) weighted using Term Frequency-Inverse Document Frequency (TFIDF). We also used character-based count vectors based on character n-grams with *n* in the range (2,5) only from the text inside word boundaries. The experiments were implemented using scikit-learn package in Python [13].

To develop deep learning models, we utilized the Bidirectional Gated Recurrent Unit (BiGRU) network and the pre-trained Bidirectional Encoder Representations from Transformers (BERT) language model [14]. The BiGRU model consists of a bidirectional GRU layer with 128 units, followed by a fully connected layer with 128 hidden units. A softmax layer is applied on top for classification with 11 units. We used 300 as input sequence length, 0.1 for dropout rate, and 20 for the number of epochs. The optimization algorithm was set to "Adam", with a categorical crossentropy loss function. For word vectors, an embedding layer was added to the beginning of the network with output dimension equal to 128. Regarding BERT-based model, we finetuned AraBERT [15] on our ArPanEmo dataset. We used the "AraBERTv0.2-Twitter-large" variant of AraBERT, with 24 encoder layers. We did not apply any text segmentation for the model. We set the learning rate to 2e-5, the batch size to 16, and the number of epochs to 10.

The results of the models indicate that our ArPanEmo dataset demonstrates promising outcomes for fine-grained emotion recognition. The BERT-based model achieves the best performance, with a weighted-average F1-score of 0.67. Furthermore, the results of the macro-average metrics for all models are relatively similar to those reported using weighted-average metrics. Macro-average metrics give equal weight to all emotion categories when computing the final average, while weighted-average metrics give more weight to larger emotion categories. Therefore, the comparable results between macro-average and weighted-average metrics suggest that the ArPanEmo dataset has a well-balanced distribution of instances for each emotion category. The results of the accuracy, macro-average and weighted-average metrics of all these models are represented in Table 4.

## Ethics statements

Our ArPanEmo dataset is completely anonymized and does not contain any personally identifiable information. For the portion of the ArPanEmo corpus obtained from Twitter, we only provide the Tweet ID and annotations because Twitter's Service Terms [16] restrict the distribution of tweet contents.

## CRediT Author Statement

**Maha Jarallah Althobaiti**: Conceptualization, Data Curation, Methodology, Visualization, Investigation, Data Validation, Writing.

## Declaration of interests

The authors declare that they have no known competing financial interests or personal relationships that could have appeared to influence the work reported in this paper.